%% file: main.tex
\documentclass[10pt,twocolumn,letterpaper]{article}

\usepackage{iccv}              


\definecolor{iccvblue}{rgb}{0.21,0.49,0.74}
\usepackage[pagebackref,breaklinks,colorlinks,allcolors=iccvblue]{hyperref}

\usepackage[accsupp]{axessibility}  

\usepackage{algorithm}
\usepackage{algorithmic}
\usepackage{graphicx}
\usepackage{booktabs}
\usepackage{framed}
\usepackage{mdframed}
\usepackage{graphicx}
\usepackage{amsmath}
\usepackage{amssymb}
\usepackage{booktabs}
\usepackage{colortbl}
\usepackage{enumerate}
\usepackage{multirow}
\usepackage{color}
\usepackage{pifont}
\usepackage{bbding}

\newcommand{\bftab}[1]{{\fontseries{b}\selectfont#1}}

\usepackage{calc}
\usepackage{epigraph}
\def\paperID{1654} 
\def\confName{ICCV}
\def\confYear{2025}

\title{Learning A Unified Template for Gait Recognition}

\def\authorBlock{
    Panjian Huang\textsuperscript{1},
    Saihui Hou\textsuperscript{1},
    Junzhou Huang\textsuperscript{2},
    Yongzhen Huang\textsuperscript{1,3}\thanks{Corresponding Author}~ \\
    \textsuperscript{1~}School of Artificial Intelligence, Beijing Normal University \\
    \textsuperscript{2~}Department of Computer Science and Engineering, The University of Texas at Arlington \\
    \textsuperscript{3~}WATRIX.AI \\
}

\begin{document}
\author{\authorBlock}
\maketitle
\input{sec/0_abstract}    
\input{sec/1_intro}
\input{sec/2_rel}
\input{sec/3_met}
\input{sec/4_exp}
\input{sec/5_con}
{
    \small
    \bibliographystyle{ieeenat_fullname}
    \bibliography{main}
}


\end{document}

%% file: sec/0_abstract.tex
\begin{abstract}
\textbf{``What I cannot create, I do not understand.''} Human wisdom reveals that creation is one of the highest forms of learning. 
For example, Diffusion Models have demonstrated remarkable semantic structure and memory in image generation, understanding, and restoration, which intuitively benefits representation learning.
However, current gait networks rarely embrace this perspective, relying primarily on learning by contrasting gait samples under varying complex conditions, leading to semantic inconsistency and uniformity issues.
To address these issues, we propose \textbf{Origins} with generative capabilities whose underlying philosophy is that \textbf{different entities are generated from a unified template}, inherently regularizing gait representations within a consistent and diverse semantic space to capture accurate gait differences.
Admittedly, learning this unified template is exceedingly challenging, as it requires the comprehensiveness of the template to encompass gait representations with various conditions.
Inspired by Diffusion Models, Origins diffuses the unified template into \textbf{timestep templates} for gait generative learning, and meanwhile transfers the unified template for gait representation learning.
Especially, gait generative and representation learning serve as a unified framework for end-to-end joint training.
Extensive experiments on CASIA-B, CCPG, SUSTech1K, Gait3D, GREW and CCGR-MINI demonstrate that Origins performs unified generative and representation learning, achieving superior performance.
\end{abstract}

%% file: sec/1_intro.tex
\section{Introduction}
\label{sec:intro}

\setlength{\epigraphwidth}{.9\columnwidth}
\renewcommand{\epigraphflush}{center}
\renewcommand{\textflush}{flushepinormal}
\epigraph{\textit{``The Tao produced One; One produced Two; Two produced Three; Three produced All things.''}}{--- \textit{Laozi, Tao Te Ching, ch.~42}}
\vspace{-0.13in}

\begin{figure}[tbp]
	\centering
	\includegraphics[width=0.99\linewidth]{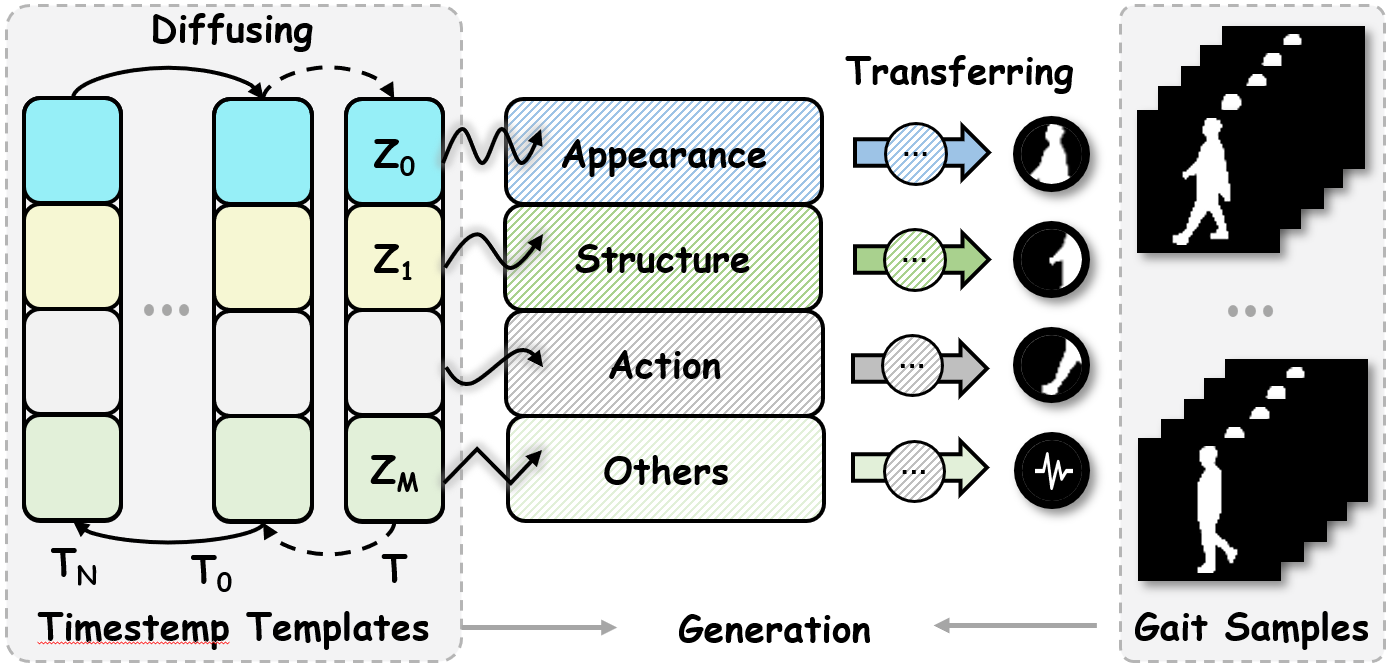}
	\caption{
    Origins performs unified generative and representation learning. The unified template $\mathcal{T}$ diffuses to the timestep templates, which generates gait representations. Meanwhile, the unified template $\mathcal{T}$ transfers to capture accurate gait differences.
	}
	\label{Template}
\end{figure}

%
According to the associative theory of creativity, individuals with higher creativity possess richer semantic structure and memory that support an expansive associative search, facilitating the combination of distant concepts into novel ideas~\cite{mednick1962associative, kenett2024role, creativity1970}.
%
Indeed, modern Generative AI (\eg, Diffusion Models) has demonstrated remarkable visual control, memory and imagination in image generation~\cite{ho2020denoising}, understanding~\cite{hudson2024soda}, and restoration~\cite{xia2023diffir}. 
%
Intuitively, a network, capable of controllably generating diverse entities across different categories, implies a powerful representation space~\cite{yu2024representation, li2023self, chen2024deconstructing, xiang2023denoising, tian2023addp}. 

\begin{figure}[tbp]
	\centering
	\includegraphics[width=0.84\linewidth]{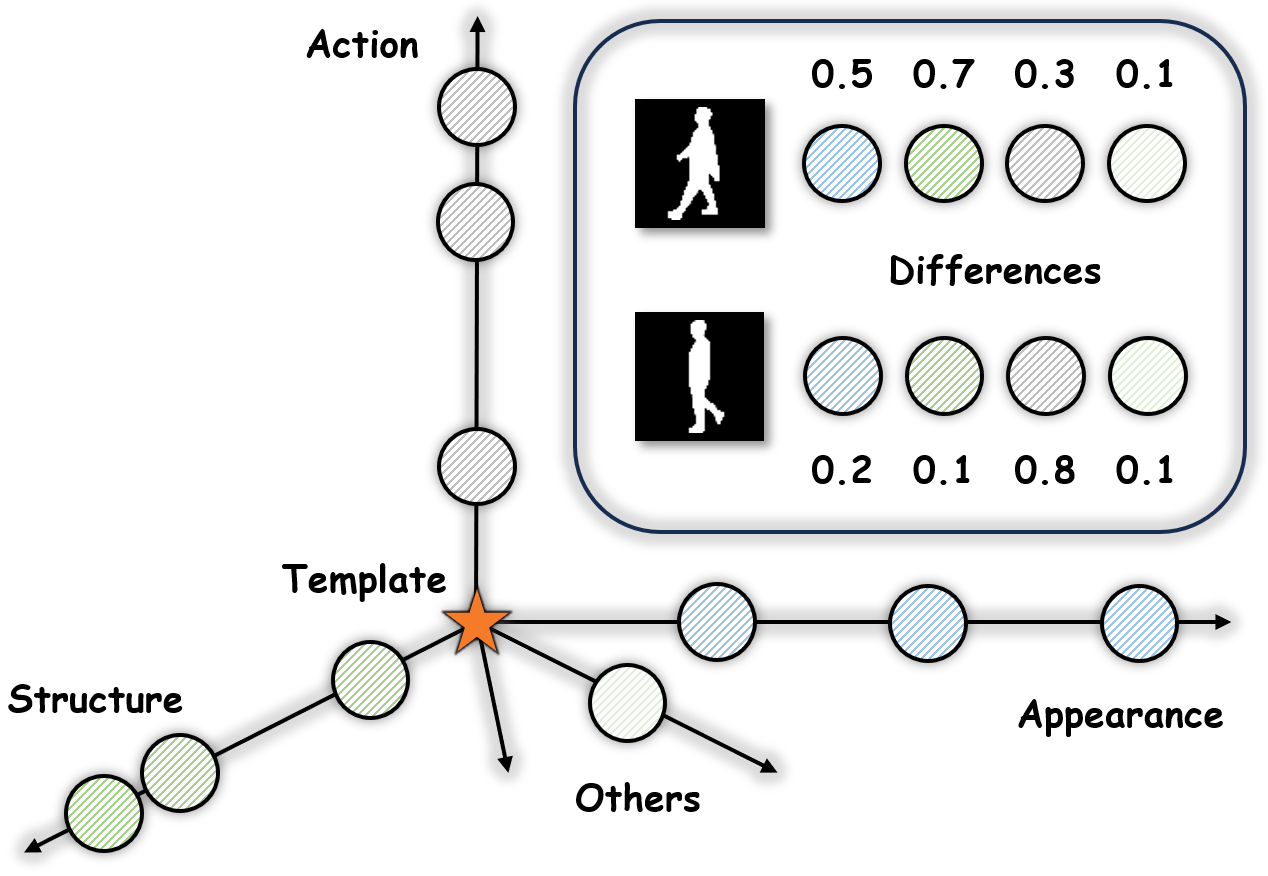}
	\caption{
    The unified template is the consistent and diverse space. The different semantic dimensions in this unified template linearly construct each gait representation.
	}
	\label{Consistent_Diverse_Space}
\end{figure}

\begin{figure*}[tbp]
	\centering
	\includegraphics[width=0.99\linewidth]{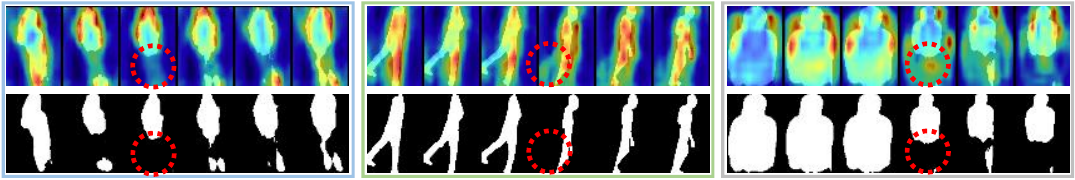}
	\caption{
    The heatmaps show that the unified template generates the missing human body parts for information supplementation and matching (\eg, the regions in the red circle of the left and right figures), and rectifies temporal information (\eg, the region in the red circle of the middle figure).
	}
	\label{Vis_Heatmap}
 
\end{figure*}

Within the scope of representation learning, gait serves as a descriptor of walking patterns for long-distance human recognition~\cite{song2022casia, sepas2022deep}. 
Although current gait recognition has made significant progress, the gait representation learning obtains identity information by contrasting gait samples under varying complex conditions (\eg, cross-view, cross-clothing, occlusion and illumination conditions), causing two problems:
\textbf{(i) Semantic Inconsistency.} Gait representations with different complex conditions may exhibit significant gaps. For example, semantic inconsistency has been observed in VPNet~\cite{ma2024learning}, where different view angles correspond to distinct prompts, indicating significant variations in the current gait representations.
Additionally, cross-view gait networks potentially learn the 3D human body projections from various 2D view angles and the chaining relations between different view angles~\cite{ma2023fine}, while cross-clothing gait networks may learn pose and motion information~\cite{huang2024integral}. Occluded gait networks may implicitly learn reasoning and filtering capabilities~\cite{huang2024occluded}. 
\textbf{(ii) Semantic Uniformity.} Facing the combination of multiple covariates, gait representations are constrained to a narrow representation space, maintaining usable patterns for all conditions. 
To address the above issues, we present Origins where \textbf{different entities are generated from a unified template}. 
The philosophy is inspired by \textbf{\textit{Tao Te Ching}}, where All things are said to originate from the ``Tao''.
Towards this goal, gait representations exhibit \textbf{Phenomena 1} (Figure~\ref{Consistent_Diverse_Space}). Gait samples with different conditions align within the semantic space. 
\textbf{Phenomena 2} (Figure~\ref{Vis_Heatmap}). The semantic space possesses sufficient diversity to encompass all gait samples with various conditions.
%

\begin{figure}[tbp]
	\centering
	\includegraphics[width=0.99\linewidth]{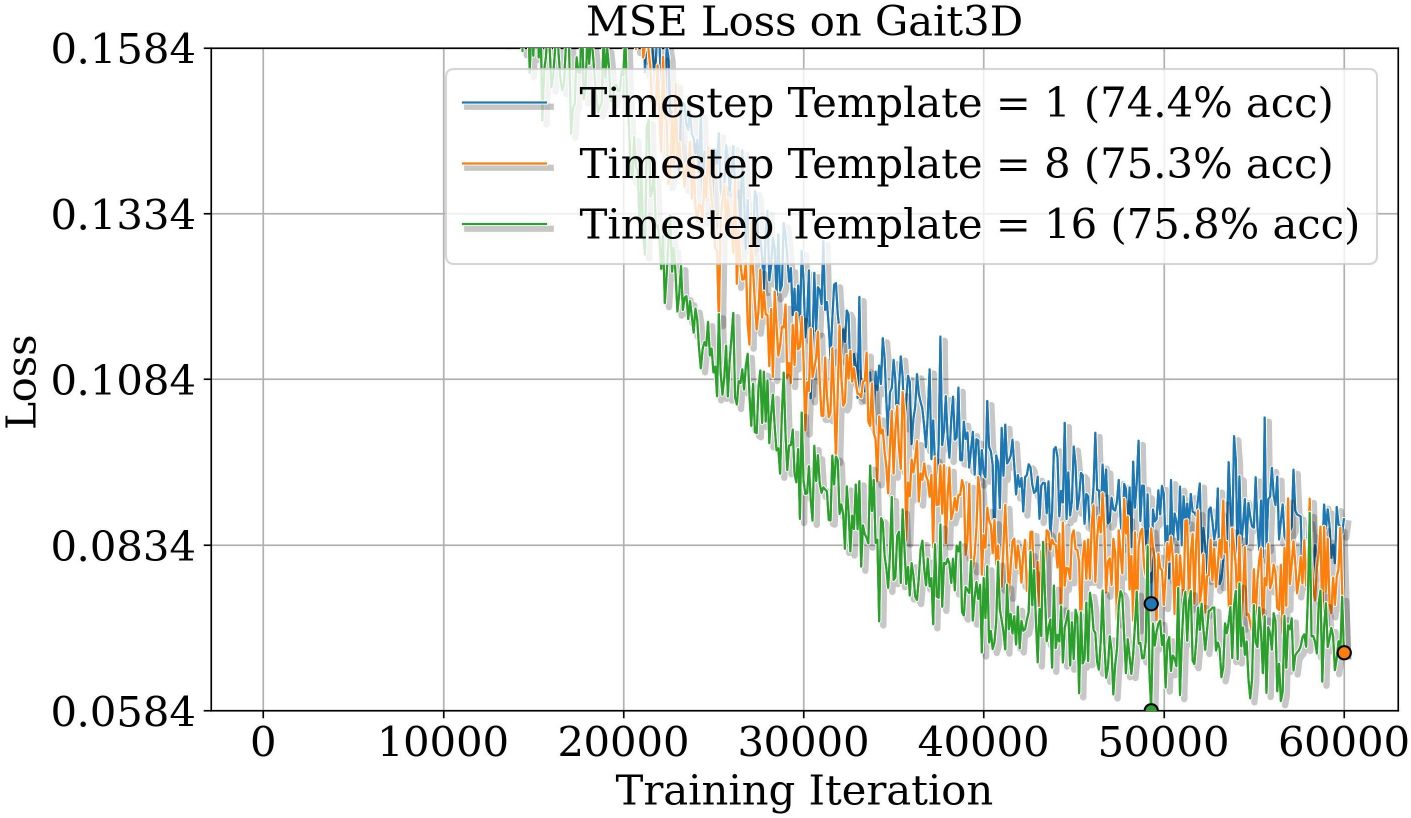}
	\caption{
    More timestep templates enable better convergence and more precise recognition performance.
	}
	\label{Loss_Curve}
\end{figure}

%
Admittedly, learning this unified template is exceedingly challenging.
Aligning and generating highly different gait samples with the template distribution encounters the convergence difficulty. 
%
As shown in Figure~\ref{Template}, Origins diffuses a unified template (\ie,  $\mathcal T$) into timestep templates (\ie, $\mathcal T_{0}, ..., \mathcal T_{\mathcal N}$), which generates gait representations, and meanwhile transfers the unified template information ($\mathcal T$) to capture accurate gait differences.
Figure~\ref{Loss_Curve} illustrates that more timestep templates enable better convergence and more precise recognition performance.
We specifically clarify relations between Origins and Diffusion Models: \\
\noindent \textbf{Inspirations.} 
Origins is inspired by diffusion models, lies in: 
(i) The step-by-step diffusion mechanism. In the vanilla diffusion training, for each sample, a random timestep is chosen, mapped through the time series and MLP, and conditioned for generation. Analogously, in the Origins training, each gait sample randomly selects a timestep template, which is transformed from the unified template through the time series and MLP, and conditioned to generate gait representations;
(ii) A consistent and diverse semantic space. The generative paradigm needs to generate every gait sample independent of cross-covariate conditions, which forces the unified template to span the entire semantic manifold.

\noindent \textbf{Highlights.} Origins aims to recognize individuals with generative capabilities. (i) No noise addition. Origins integrates the generative process to learn a consistent yet diverse semantic space without sampling new gait samples from a known noise distribution (\eg, Gaussian distribution). 
(ii) Implicit timestep template relations. 
Origins does not follow a conventional diffusion model where timestep relations are explicitly equivalent to the Markov chain. Instead, it implicitly learns the timestep template relations by randomly sampling to generate gait representations.
%
(iii) No identity constraint.
The generative evolution aims to learn a consistent and diverse space (\ie, a unified template), independent of identity retention, where Origins only adopt the unified template and real samples for identification.

%
Our main contributions can be summarized as follows: 
\begin{itemize}[left=5mm]
     \item We propose a novel framework Origins where the unified generative and representation learning regularizes gait representations within a consistent and diverse semantic space, addressing the semantic inconsistency and uniformity issues.
    \item We design Diffusing Timestep Templates to alleviate the convergence difficulty, and Transferring Unified Template to capture accurate gait differences. 
    \item We evaluate Origins on six public benchmarks, CASIA-B, CCPG, SUSTech1K, Gait3D, GREW and CCGR-MINI, demonstrating the effectiveness and achieving superior performance.
\end{itemize}

%% file: sec/2_rel.tex
\section{Related Work}
\label{sec:Rel}

\subsection{Diffusion Models and Representation Learning}
\label{sec:DMs}

\noindent \textbf{Diffusion Models (DMs).} Diffusion models gradually add noise to input data and learn to reverse this process for data generation, making them a probabilistic generative framework. (i) Architectures. DDPMs~\cite{ho2020denoising} leverages a U-Net architecture for denoising, which enables effective noise prediction across varying timesteps. LDMs~\cite{rombach2022high} apply diffusion processes in the latent space of a pre-trained variational autoencoder. This innovation significantly reduces computational overhead while maintaining high generation quality. DiT~\cite{peebles2023scalable} adapts Vision Transformers (ViTs) by tokenizing inputs and attention-based mechanisms for more flexibility.
(ii) Guidance Techniques. Classifier Guidance~\cite{dhariwal2021diffusion} uses gradients from pre-trained noise-robust classifiers to guide the diffusion process. Classifier-Free Guidance~\cite{ho2022classifier} avoids reliance on external classifiers by training a diffusion model with both conditional and unconditional inputs. Self-Guided Diffusion~\cite{hu2023self} introduces a self-supervised framework that generates guidance signals via clustering.

\noindent \textbf{Representation Learning Based on DMs.}
Representation learning based on DMs can be broadly categorized into five types.
(i) Leveraging Intermediate Activations. DDPM-Seg~\cite{baranchuk2021label} extracts intermediate feature activations from decoder blocks of DDPMs for semantic segmentation.
(ii) Knowledge Transfer. RepFusion~\cite{yang2023diffusion} employs reinforcement learning to dynamically extract representations from diffusion models, which are distilled into student networks for downstream tasks.
(iii) Reconstructing Diffusion Models. l-DAE~\cite{chen2024deconstructing} reconstructs DDPMs into autoencoder for self-supervised learning, highlighting the role of denoising in representation learning. 
(iv) Generative Augmentation. GAM~\cite{ayromlou2024can} employs latent diffusion models to create augmented views of training data, enhancing the generalization of learned representations across diverse datasets.
(v) Joint Diffusion Models. HybirdViT~\cite{yang2022your} and ADDP~\cite{tian2023addp} combine generative and discriminative objectives in a single model, improving performances in both.
Origins aims to construct a consistent and diverse gait representation space with generative capabilities, which falls under this scope.

\subsection{Gait Recognition}
\label{sec:GR}

\noindent \textbf{Model-Based Gait Recognition.} These methods focus on human structure representations.
%
PoseGait~\cite{liao2020model} uses 3D human body pose features, including joint angles, limb lengths, and motion patterns, to enhance gait robustness.
GaitGraph~\cite{teepe2021gaitgraph} and GaitGraph2~\cite{teepe2022towards} employ Graph Convolutional Networks (GCNs) to model robust spatio-temporal information.
GaitTR~\cite{zhang2023spatial} and GaitMixer~\cite{pinyoanuntapong2023gaitmixer} incorporate self-attention mechanisms to capture long-range spatial correlations.
GPGait~\cite{fu2023gpgait} proposes a generalized pose-based gait framework, improving cross-domain generalization by transforming pose data into a unified representation.
SMPLGait~\cite{zheng2022gait3d} introduces the SMPL model to integrate dense 3D mesh representations.
SkeletonGait~\cite{fan2024skeletongait}, HiH~\cite{wang2023hih}, and GaitHeat~\cite{fu2024cut} introduce gaussian-approximated skeleton maps for structural analysis and shape details.

\noindent \textbf{Appearance-Based Gait Recognition.} These methods primarily employ human shape representations. 
GaitSet~\cite{chao2019gaitset} proposes a set-based method with a flexible, permutation-invariant framework.
GaitPart~\cite{fan2020gaitpart} introduces a temporal part-based framework with fine-grained body-part motion.
GaitGL~\cite{lin2021gait} combines global and local features with 3D CNNs to capture fine-grained temporal-spatial patterns.
GaitBase~\cite{fan2023opengait} proposes a simple yet robust baseline for the real-world applications.
DANet~\cite{ma2023dynamic}, DyGait~\cite{wang2023dygait}, HSTL~\cite{wang2023hierarchical}, VPNet~\cite{ma2024learning}, GLGait~\cite{peng2024glgait} and GaitMoE~\cite{huang2024occluded} focus on dynamic local-global gait representations.
GaitGCI~\cite{dou2023gaitgci}, GaitCSV~\cite{wang2023causal}, CLTD~\cite{xiong2024causality}, GaitC$^3$I~\cite{wang2025gaitc}, QAGait~\cite{wang2024qagait}, Free Lunch~\cite{wang2024free} and GaitAttack~\cite{hou2025edge} address confounders and noises with the interpretability.
In addition, there are some research with other gait modalities, such as GaitEdge~\cite{liang2022gaitedge} with RGB data; GaitParsing~\cite{wang2023gaitparsing}, LandmarkGait~\cite{wang2023landmarkgait} and ParsingGait~\cite{zheng2023parsing} with parsing information; LidarGait~\cite{shen2023lidargait} with point clouds; MMGaitFormer~\cite{cui2023multi} and CL-Gait~\cite{guo2024camera} with multimodal data.

\begin{figure*}[tbp]
	\centering
	\includegraphics[width=1.0\linewidth]{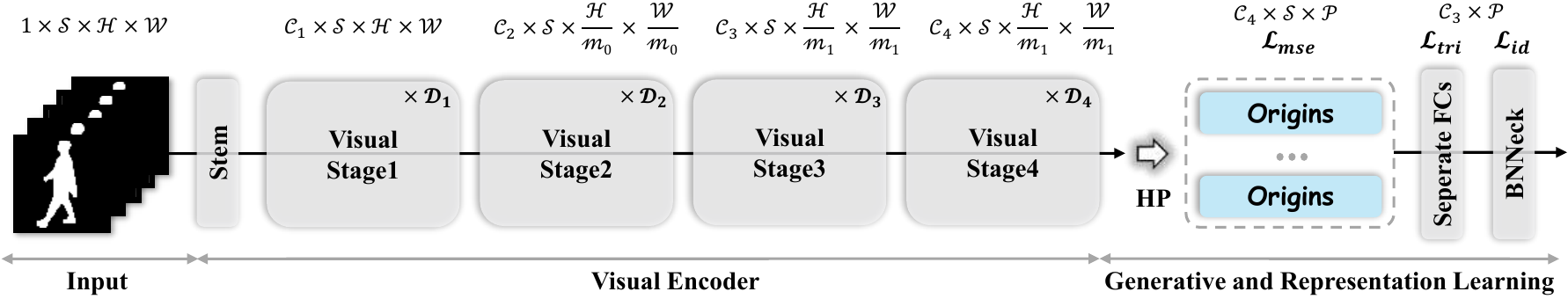}
	\caption{
    The gait network with Origins. HP represents Horizontal Partition, and Visual Stage consists of basic convolution blocks. After the gait silhouette sequence passes through Visual Encoder and HP, the part representations are fed into the respective Origins and Recognition Head for generative and representation learning.
	}
	\label{Origins}
 
\end{figure*}

%% file: sec/3_met.tex
\section{Methodology}

\begin{figure}[tbp]
	\centering
	\includegraphics[width=0.99\linewidth]{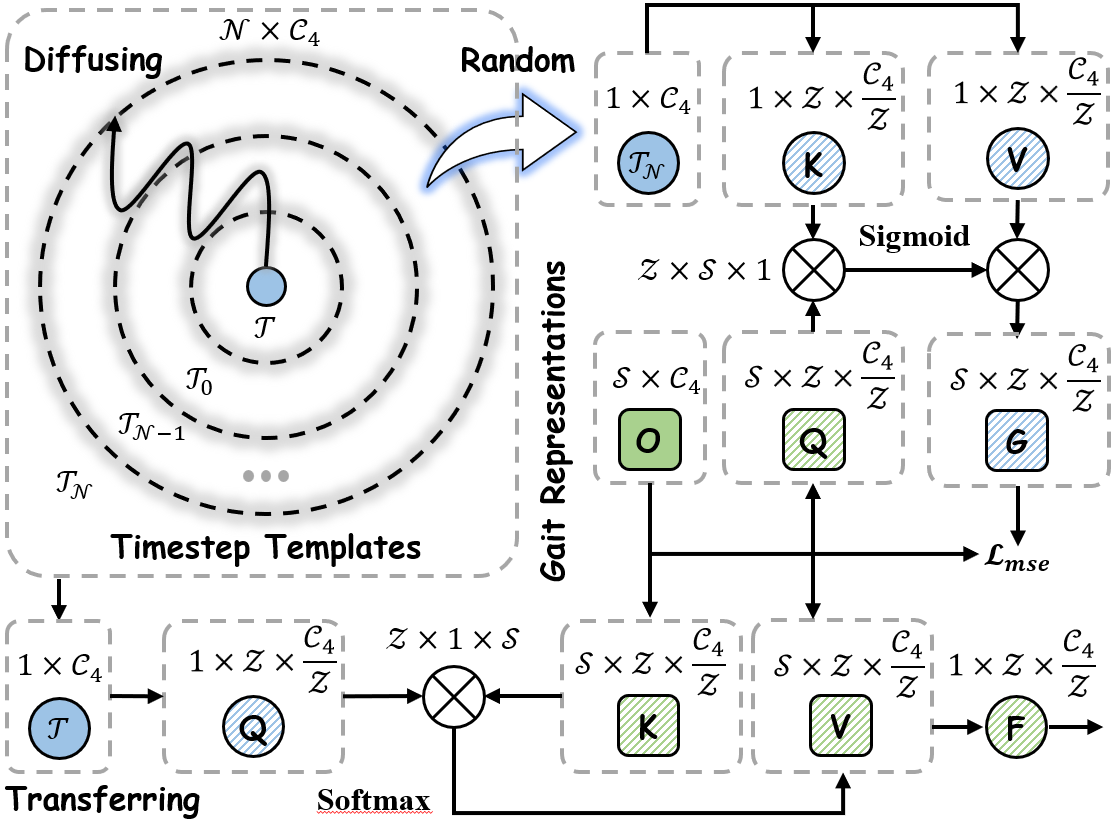}
	\caption{
    The overview of Origins, best viewed in colors and shapes. The $\mathcal N$, $\mathcal C_{4}$ and $\mathcal Z$ denote the number, dimension and multi-heads of timestep templates. The unified template $\mathcal{T}$ diffuses timestep templates by time serialization for generation and transfers information to gait representations for recognition.
	}
	\label{Origins_Detail}
\end{figure}

%

In this section, we first introduce the gait network with Origins in Sec.~\ref{sec:overview}, then present the overview of Origins in Sec.~\ref{sec:origins}, and finally, describe the unified gait generative and representation learning as a unified framework for end-to-end joint training in Sec.~\ref{sec:joint}.

\subsection{Overview}
\label{sec:overview}
As shown in Figure~\ref{Origins}, the vanilla gait framework typically consists of a Visual Encoder ($\mathcal E$), a Horizontal Partition ($\mathcal H \mathcal P$), a Recognition Head ($\mathcal R \mathcal H$), and a Joint Loss ($\mathcal L$). In this work, the Visual Encoder employs a Stem and several Visual Stages (as in ResNet~\cite{he2016deep}). Notably, we introduce the general backbone rather than sophisticated designs for better validating the effectiveness of generative learning. The $\mathcal H \mathcal P$ horizontally splits the human body ($\mathcal P$), extracting and matching finer-grained identity information. Finally, the $\mathcal R \mathcal H$ performs feature mapping for optimization with Joint Loss (\ie, Triplet Loss~\cite{hermans2017defense} and Cross-Entropy Loss~\cite{fan2023opengait}). 
The proposed Origins is embedded after the $\mathcal H \mathcal P$ and individually on each human part, regularizing the part representations into a consistent and diverse semantic space. 
\textbf{\textit{Here, we omit the part index for simplicity.}}
Formally, given the gait silhouette sequence $\mathcal X \in \mathbb{R}^{1 \times \mathcal S \times \mathcal H \times \mathcal W}$, where $1, \mathcal S, \mathcal H, \mathcal W$ represent channel, consecutive $\mathcal S$ frames, height and width dimensions, the process of Visual Encoder and Horizontal Partition is formulated as follows:
\begin{equation}
\begin{aligned}
\mathcal O = \mathcal H \mathcal P (\mathcal E (\mathcal X))
\end{aligned}
\end{equation}
where $\mathcal O \in \mathbb{R}^{\mathcal C_{4} \times \mathcal S \times \mathcal P}$ is the part representations. 
Similar to~\cite{rombach2022high}, generative learning in the latent space is generally more efficient and flexible than in the pixel space.

\subsection{Origins}
\label{sec:origins}
Existing gait paradigms primarily rely on learning the invariant gait representations by contrasting gait samples with different conditions. However, the complex real-world environment (\eg, cross-view and cross-clothing conditions) inevitably causes semantic inconsistency and uniformity (\eg, the view angle chaining relations vs. the motion information).
%
\begin{table*}[tbp]
\centering
	\caption{
        DataBase and Architectures. Id. and Seq. denote the number of identities and sequences. CV, BG and CL refer to cross-view and carrying bags and cross-clothing conditions. $\mathcal{D}$ and $\mathcal{C}$ denote the number of conv blocks and the channels in each visual stage.
	}
 \setlength\tabcolsep{5.0pt}
 \resizebox{\linewidth}{!}{
\begin{tabular}{c|c|cc|cc|c|c|c|c}
\hline
\multirow{2}{*}{Environment} & \multirow{2}{*}{Dataset} & \multicolumn{2}{c|}{Train}            & \multicolumn{2}{c|}{Test}           & \multirow{2}{*}{Condition} & Stage                & Channels                & \multirow{2}{*}{Strides} \\ \cline{3-6} \cline{8-9}
                             &                          & \multicolumn{1}{c|}{Id.}    & Seq.    & \multicolumn{1}{c|}{Id.}   & Seq.   &                            & {[}$\mathcal D_{1}$, $\mathcal D_{2}$, $\mathcal D_{3}$, $\mathcal D_{4}${]} & {[}$\mathcal C_{1}$, $\mathcal C_{2}$, $\mathcal C_{3}$, $\mathcal C_{4}${]}    &                          \\ \hline
\multirow{2}{*}{Constrained} & CASIA-B~\cite{yu2006framework}                  & \multicolumn{1}{c|}{74}     & 8,140   & \multicolumn{1}{c|}{50}    & 5,500  & CV, BG, CL           & {[}1, 1, 1. -{]}     & {[}64, 128, 256, -{]}   & {[}1, 2, 1, -{]}         \\ \cline{2-10} 
                             & CCPG~\cite{li2023depth}                     & \multicolumn{1}{c|}{100}    & 8,187   & \multicolumn{1}{c|}{100}   & 8,095  & CV, BG, CL           & {[}1, 1, 1, 1{]}     & {[}64, 128, 256, 512{]}   & {[}1, 2, 2, 1{]}         \\ \hline
\multirow{4}{*}{In-the-wild} & SUSTech1K~\cite{shen2023lidargait}                   & \multicolumn{1}{c|}{200}  & 5988  & \multicolumn{1}{c|}{850} & 19,228  & Real-world                 & {[}1, 1, 1, 1{]}     & {[}64, 128, 256, 512{]} & {[}1, 2, 2, 1{]}         \\ \cline{2-10} 

& CCGR-MINI~\cite{zou2024cross} & \multicolumn{1}{c|}{571}  & 27507  & \multicolumn{1}{c|}{399} & 20377  & Real-world                 & {[}1, 4, 4, 1{]}     & {[}64, 128, 256, 512{]} & {[}1, 2, 2, 1{]}         \\ \cline{2-10}

& Gait3D~\cite{zheng2022gait3d}                   & \multicolumn{1}{c|}{3,000}  & 18,940  & \multicolumn{1}{c|}{1,000} & 6,369  & Real-world                 & {[}1, 4, 4, 1{]}     & {[}64, 128, 256, 512{]} & {[}1, 2, 2, 1{]}         \\ \cline{2-10} 

                             & GREW~\cite{zhu2021gait}                     & \multicolumn{1}{c|}{20,000} & 102,887 & \multicolumn{1}{c|}{6,000} & 24,000 & Real-world                 & {[}3, 4, 6, 3{]}     & {[}64, 128, 256, 512{]} & {[}1, 2, 2, 1{]}         \\ \hline
\end{tabular}
}
\label{datasets}
\end{table*}
To this end, we propose Origins with the generative capability to regularize gait representations within a consistent and diverse semantic space to capture accurate gait differences. As shown in Figure.~\ref{Origins_Detail}, Origins presents the semantic consistency as prior, applying a unified template to generate gait representations for the entire gait database, implying that each gait sequence can be constructed within the space spanned by this unified template.

\noindent \textbf{Diffusing Timestep Templates.} 
Admittedly, learning this unified template is extremely challenging, as it requires the comprehensiveness of the template to encompass gait representations with various conditions, facing the convergence difficulty as many generative models (\eg, GANs~\cite{goodfellow2020generative} and VQVAEs~\cite{van2017neural}). 
Inspired by the step-by-step mechanism in Diffusion Models~\cite{rombach2022high, nichol2021glide, dhariwal2021diffusion}, Origins diffuses this unified template into timestep templates, where each is derived from the unified template through time series modeling and MLP. Formally, Given the learnable unified template $\mathcal T \in \mathbb{R}^{1 \times \mathcal C_{4}}$, the diffusion process for timestep templates is as follows:
\begin{equation}
\mathcal T_{\mathcal N} \;=\;
\mathcal T + \operatorname{MLP}\!\bigl(t_{\mathcal N}\bigr)
\end{equation}
\begin{equation}
\begin{aligned}
t_{\mathcal N} = \bigl[
  \cos(\omega_1 \mathcal N),\,\dots,\,
  \cos(\omega_{\tfrac{\mathcal C_{4}}{2}}\mathcal N),\\
  \qquad
  \sin(\omega_1 \mathcal N),\,\dots,\,
  \sin(\omega_{\tfrac{\mathcal C_{4}}{2}}\mathcal N)
\bigr]
\end{aligned}
\end{equation}
\begin{equation}
\omega_k = 10000^{-\frac{2(k-1)}{\mathcal C_{4}}}, 
\qquad k = 1,\dots,\tfrac{\mathcal C_{4}}{2}
\end{equation}
where $\mathcal T_{\mathcal N} \in\mathbb{R}^{1 \times \mathcal C_{4}}$, $\mathcal N$ is the scalar timestep, $\omega_k$ defines a frequency schedule.
Similar to the training stage in Diffusion Models, Origins does not require generation through all sequential timestep templates. Instead, it implicitly learns the timestep template relations by randomly sampling to generate gait representations.
Given a gait representation sequence $\mathcal O \in \mathbb{R}^{\mathcal S \times \mathcal C_{4}}$ and a \textbf{\textit{Random}} sampled timestep template $\mathcal T_{\mathcal N} \in \mathbb{R}^{1 \times \mathcal C_{4}}$, the generation process is as follows:
\begin{equation}
\begin{aligned}
\mathcal Q  = \mathcal W ^{\mathcal Q} \mathcal O,\quad \mathcal K = \mathcal W ^{\mathcal K} \mathcal T_{\mathcal N},\quad \mathcal V = \mathcal W ^{\mathcal V} \mathcal T_{\mathcal N}
\end{aligned}
\end{equation}
\begin{equation}
\begin{aligned}
\mathcal G = \operatorname{Sigmoid} (\mathcal Q \otimes \mathcal K^{T}) \otimes \mathcal V
\end{aligned}
\end{equation}
where $\mathcal W ^{\mathcal Q}$, $\mathcal W ^{\mathcal K}$ and $\mathcal W ^{\mathcal V} \in \mathbb{R}^{\mathcal C_{4} \times \mathcal Z \times \frac{\mathcal C_{4}}{\mathcal Z}}$ are mapping functions and $\mathcal Z$ denotes the number of multi-heads in the attention mechanism, similar to Transformers~\cite{vaswani2017attention}.
Each head denotes one type of semantic information.
This generation process is optimized by MSE Loss:
\begin{equation}
\begin{aligned}
\mathcal L_{mse} = \frac{1}{S} (\mathcal G - \operatorname{detach}(\mathcal O))^2
\end{aligned}
\end{equation}
The generated gait representation $\mathcal G$ is composed of $\mathcal Z$ multiple heads of the timestep template ($\mathcal Z$ is empirically set to 16). 
Diffusing Timestep Templates regularizes all gait representations into a consistent and diverse space.

\noindent \textbf{Transferring Unified Template.}
To further exploit template information for capturing accurate gait differences, Origins transfers the unified template $\mathcal T$ to compress a gait sequence into one token. 
Specifically, as Diffusing Timestep Template progresses, the unified template gradually accumulates rich gait knowledge from the entire gait database. Therefore, the unified template enables to compresses each gait sequence as completely and effectively as possible, which is inspired by the idea of ``Compression as Intelligence''~\cite{schmidhuber2008driven}.
The process is as follows:
\begin{equation}
\begin{aligned}
\mathcal Q  = \mathcal W ^{\mathcal Q} \mathcal T,\quad \mathcal K = \mathcal W ^{\mathcal K} \mathcal O,\quad \mathcal V = \mathcal W ^{\mathcal V} \mathcal O
\end{aligned}
\end{equation}
\begin{equation}
\begin{aligned}
\mathcal F = \operatorname{Softmax} (\mathcal Q \otimes \mathcal K^{T}) \otimes \mathcal V
\end{aligned}
\end{equation}
where $\mathcal W ^{\mathcal Q}$, $\mathcal W ^{\mathcal K}$ and $\mathcal W ^{\mathcal V} \in \mathbb{R}^{\mathcal C_{4} \times \mathcal Z \times \frac{\mathcal C_{4}}{\mathcal Z}}$ are mapping functions and $\mathcal Z$ denotes the number of multi-heads in the attention mechanism.
Finally, this one token $\mathcal F$ is fed into the following Recognition Head.

\noindent \textbf{Summarize}. Origins includes that (i) Only the unified template serves as ($\mathcal Q$) transfers to extract discriminative representations from real gait samples ($\mathcal K$, $\mathcal V$); (ii) Meanwhile, the unified template diffuses into timestep templates; (iii) Only timestep templates serve as bases ($\mathcal K$, $\mathcal V$) conditioned by real gait samples ($\mathcal Q$) for generative learning, where each gait sample randomly selects a timestep template.

\subsection{Training Details}
\label{sec:joint}
Origins unifies gait generative and representation learning for end-to-end joint training.
The joint loss is as follows:
\begin{equation}
\begin{aligned}
\mathcal L = \mathcal L _{mse} + \mathcal L _{tp} + \mathcal L _{ce}
\end{aligned}
\end{equation}
where $\mathcal L_{mse}$, $\mathcal L _{tp}$ and $\mathcal L _{ce}$ denote MSE Loss, Triplet Loss~\cite{hermans2017defense} and Cross Entropy Loss.

%% file: sec/4_exp.tex
\section{Experiments}
\label{sec:Exp}

\begin{table*}[t]
\setlength{\tabcolsep}{4.0pt}
\centering
\footnotesize
 	\caption{
       The performance comparisons on CCPG under various conditions, including full-body cloth changes (CL), upper-body cloth changes (UP), lower-body cloth changes (DN), and backpacks changes (BG), reported with Rank-1 accuracy (\%).
	}
        \setlength\tabcolsep{6pt}
		\resizebox{0.9999\linewidth}{!}{%
			\begin{tabular}{c|c|c|ccccc|ccccc}
				\hline
				\multirow{2}{*}{Paradigm} & \multirow{2}{*}{Method} & \multirow{2}{*}{Venue} & \multicolumn{5}{c}{Gait Evaluation Protocol} & \multicolumn{5}{c}{ReID Evaluation Protocol}\\
				\cline{4-13}
				& & &CL &UP &DN &BG &Mean &CL &UP &DN &BG &Mean \\
				\hline
                \multirow{4}{*}{Model} &GaitGraph2~\cite{teepe2021gaitgraph} & CVPRW22 &5.0 &5.3 &5.8 &6.2 &5.1   &5.0 &5.7 &7.3 &8.8 &6.7\\
				& Gait-TR~\cite{zhang2023spatial} &  ES23 &15.7 &18.3 &18.5 &17.5 &17.5 &24.3 &28.7 &31.1 &28.1 &28.1\\
				& MSGG~\cite{peng2024learning} &   MTA23 &29.0 &34.5 &37.1 &33.3 &33.5 &43.1 &52.9 &57.4 &49.9 &50.8\\
                & SkeletonGait~\cite{fan2024skeletongait} & AAAI24 &40.4 &48.5 &53.0 &61.7 &50.9 &52.4 &65.4 &72.8 &80.9 &67.9\\
                \hline
				\multirow{5}{*}{Appearance} & GaitSet~\cite{chao2019gaitset} & AAAI19 &60.2 &65.2 &65.1 &68.5 &64.8   &77.5 &85.0 &82.9 &87.5 &83.2\\
				& GaitPart~\cite{fan2020gaitpart} & CVPR20 &64.3 &67.8 &68.6 &71.7 &68.1 &79.2 &85.3 &86.5 &88.0 &84.8\\
                & OGBase~\cite{li2023depth} & CVPR23 &52.1 &57.3 &60.1 &63.3 &58.2 &70.2 &76.9 &80.4 &83.4 &77.7\\
				& GaitBase~\cite{fan2023opengait} & CVPR23 &71.6 &75.0 &76.8 &78.6 &75.5 &88.5 &92.7 &93.4 &93.2 &92.0\\ 
				& DeepGaitV2~\cite{11023027} & TPAMI25 &78.6 &84.8 &80.7 &89.2 &83.3 &90.5 &96.3 &91.4 &96.7 &93.7 \\ 
               \rowcolor{blue!8} & Origins-S (\bftab{ours})  &- &\bftab{84.3} &\bftab{90.2} &\bftab{86.4} &\bftab{93.6}&\bftab{88.6} &\bftab{93.4} &\bftab{97.6} &\bftab{94.6} &\bftab{97.6} &\bftab{95.8}\\
                \hline
			\end{tabular}
			
		}
	\label{ccpg}
\end{table*}

\begin{table*}[t]
\setlength{\tabcolsep}{4.0pt}
\centering
\footnotesize
 	\caption{
       The performance comparisons on SUSTech1K are reported with Rank-1 and Rank-5 accuracy (\%).
	}
        \setlength\tabcolsep{2pt}
		\resizebox{0.9999\linewidth}{!}{%
			\begin{tabular}{c|c|c|cccccccc|cc}
				\hline
				\multirow{2}{*}{Paradigm} & \multirow{2}{*}{Method} & \multirow{2}{*}{Venue} & \multicolumn{8}{c}{Probe Sequence (Rank-1)} & \multicolumn{2}{c}{Overall}\\
				\cline{4-13}
				& & &Normal &Bag &Clothing &Carrying &Umbrella &Uniform &Occlusion &Night &Rank-1 &Rank-5 \\
				\hline
                \multirow{4}{*}{Model} &GaitGraph2~\cite{teepe2021gaitgraph} & CVPRW22 &22.2 &18.2 &6.8 &18.6 &13.4 &19.2 &27.3 &16.4 &18.6 &40.2\\
				& Gait-TR~\cite{zhang2023spatial} &  ES23 &33.3 &31.5 &21.0 &30.4 &22.7 &34.6 &44.9 &23.5 &30.8 &56.0\\
				& MSGG~\cite{peng2024learning} &   MTA23 &67.11 &66.16 &35.92 &63.31 &61.58 &58.07 &66.59 &17.88 &33.8 &-\\
                & SkeletonGait~\cite{fan2024skeletongait} & AAAI24 &67.9 &63.5 &36.5 &61.6 &58.1 &67.2 &79.1 &50.1 &63.0 &83.5\\
                \hline
				\multirow{5}{*}{Appearance} &GaitSet~\cite{chao2019gaitset} & AAAI19 &69.1 &68.2 &37.4 &65.0 &63.1 &61.0 &67.2 &23.0 &65.0 &84.8\\
				&GaitPart~\cite{fan2020gaitpart} & CVPR20 &62.2 &62.8 &33.1 &59.5 &57.2 &54.8 &57.2 &21.7 &59.2 &80.8\\
                &GaitGL~\cite{lin2021gait} & ICCV21  &67.1 &66.2 &35.9 &63.3 &61.6 &58.1 &66.6 &17.9 &63.1 &82.8\\
				&GaitBase~\cite{fan2023opengait} & CVPR23 &81.5 &77.5 &49.6 &75.8 &75.5 &76.7 &81.4 &25.9 &76.1 &89.4 \\
                & DeepGaitV2~\cite{11023027} & TPAMI25 &87.4 &84.1 &53.4 &81.3 &86.1 &84.8 &88.5 &28.8 &82.3 &92.5 \\
                \rowcolor{blue!8} & Origins-S (\bftab{ours})  &- &\bftab{91.4} &\bftab{88.1} &\bftab{64.8} &\bftab{86.0}&\bftab{89.8} &\bftab{88.9} &\bftab{92.8} &\bftab{29.6} &\bftab{86.9} &\bftab{94.2}\\
				\hline
			\end{tabular}
			
		}
	\label{SUSTech1K}
\end{table*}

\subsection{Datasets}
Gait datasets are generally divided into two subsets: constrained and in-the-wild, based on their collection environments. As shown in Table.~\ref{datasets}, constrained datasets (\ie, CASIA-B~\cite{yu2006framework} and CCPG~\cite{li2023depth}) typically provide quantified conditional benchmarks but include a relatively small number of identities, whereas in-the-wild datasets (\ie, SUSTech1K~\cite{shen2023lidargait}, Gait3D~\cite{zheng2022gait3d}, GREW~\cite{zhu2021gait}) and CCGR-MINI~\cite{zou2024cross} involve more complex environments and a large number of identities. Origins is thoroughly evaluated the effectiveness across these widely-used gait benchmarks.

\noindent \textbf{CASIA-B} contains 124 subjects with 11 camera views and 3 scenarios:  normal walking (NM), carrying a bag (BG) and cloth-changing condition (CL).

\noindent \textbf{CCPG} provides the challenge of rich cross-clothing conditions and contains 200 subjects with over 16,000 sequences.

\noindent \textbf{SUSTech1K} consists of 1,050 subjects under various real-world conditions such as Clothing, Occlusion and Night.

\noindent \textbf{Gait3D} collects gait data in a supermarket, addressing practical gait recognition. it includes 3,000 subjects and 25,309 sequences, divided into a training set of 2,000 subjects and a testing set of 1,000 subjects.

\noindent \textbf{GREW} is a large-scale gait dataset in the wild, including 26,345 subjects and 128,671 sequences recorded by 882 cameras. The benchmark consists of 20,000 subjects for training and 6,000 for testing.

\noindent \textbf{CCGR-MINI} is provided by the CCGR~\cite{zou2024cross} team, which serves as an alternative to CCGR, offers fast training, and maintains equivalent covariates.
\subsection{Implementation Details}
The details of the training process are as follows. 
\textbf{Inputs.} Each input gait sequence consists of 30 frames, and all silhouettes are resized to 64 × 44. The batch size {]}$\mathcal I$, $\mathcal J${]} is consistent with \cite{fan2023opengait}, where $\mathcal I$ represents the number of subjects sampled per mini-batch, and $\mathcal J$ represents the number of sequences sampled per subject.
\textbf{Networks.} As shown in Table.~\ref{datasets}, we provide Origins-T, Origins-S, Origins-M, and Origins-L based on network depths. Origins-T consists of 3 Bottleneck3D Stages with block numbers [1, 1, 1] and channels [64, 128, 256]. Origins-S comprises 4 Bottleneck3D Stages with block numbers [1, 1, 1, 1] and channels [64, 128, 256, 512]. Origins-M consists of 1 Basic2D Stage and 3 Pseudo3D Stages, with block numbers [1, 4, 4, 1] and channels [64, 128, 256, 512]. Origins-L is composed of 4 Bottleneck3D Stages, with block numbers [3, 4, 6, 3] and channels [64, 128, 256, 512].
\textbf{Optimization.} We use the optimizer of SGD with an initial learning rate of 0.1, which is decreasing by a factor of 0.1 per $[20K, 40K, 50K]$, $[20K, 40K, 50K]$, $[20K, 30K, 40K]$, $[20K, 40K, 50K]$, $[80K, 120K, 150K]$, $[30K, 55K, 65K]$ for CASIA-B (Total 60$K$), CCPG (Total 60$K$), SUSTech1K (Total 50$K$), Gait3D (Total 60$K$), GREW (Total 180$K$) and CCGR-MINI (Total 80$K$). 
All the models are trained on NVIDIA 8×3090 GPUs.

\subsection{Results on Constrained Scenario}
We first validate the effectiveness of Origins across various complex scenarios on CASIA-B and CCPG.

\noindent \textbf{CASIA-B.} As shown in Table.~\ref{casiab}, Origins-T achieves state-of-the-art (SoTA) performance on all scenarios, with an average accuracy of 95.7\%, demonstrating that the unified template possesses a consistent and diverse semantic space to improve the individual distinctiveness and discriminability of gait representations under different conditions.

\noindent \textbf{CCPG.} As shown in Table.~\ref{ccpg}, Origins-S achieves outstanding performance (\eg, 88.6\% in Gait Evaluation Protocol and 95.8\% in ReID Evaluation Protocol) in more challenging clothing-change scenarios (\eg, full-body, upper-body, lower-body, and backpacks changes), which indicates the ability to capture accurate gait differences, significantly alleviating one of the biggest bottlenecks in current gait recognition: the clothing-change problem.

\begin{table}[t]
\centering
\footnotesize
 	\caption{
        The performance comparisons on CASIA-B are reported with Rank-1 accuracy (\%).
	}
        \setlength\tabcolsep{5pt}
		\resizebox{0.9999\linewidth}{!}{%
			\begin{tabular}{c|c|c|c|c|c}
				\hline
				{Method} & {Venue} & {NM} &{BG} &{CL} &{Mean}\\
				\hline
				GaitSet~\cite{chao2019gaitset} & AAAI19 &95.0 &87.2 &70.4 &84.2\\
				GaitPart~\cite{fan2020gaitpart} & CVPR20 &96.2 &91.5 &78.7 &88.8\\
                GLN~\cite{hou2020gait} &ECCV20 &96.9 &94.0 &77.5 &89.5\\
				GaitGL~\cite{lin2021gait} & ICCV21 &97.4 &94.5 &83.6 &91.8\\
                QAGait~\cite{wang2024qagait} & AAAI24 &97.9 &94.6 &78.2 &90.2\\
				GaitBase~\cite{fan2023opengait} & CVPR23 &97.6 &94.0 &77.4 &89.8\\
				DANet~\cite{ma2023dynamic} & CVPR23 &98.0 &95.9 &89.9 &94.6\\
				GaitGCI~\cite{dou2023gaitgci} & CVPR23 &97.9 &95.0 &86.4 &93.1\\
				DyGait~\cite{wang2023dygait} & ICCV23 &98.4 &96.2 &87.8 &94.1\\
				HSTL~\cite{wang2023hierarchical} & ICCV23 &98.1 &95.9 &88.9 &94.3\\
                VPNet~\cite{ma2024learning} & CVPR24 &98.3 &96.3 &90.0 &94.9\\
                DeepGaitV2~\cite{11023027} & TPAMI25 &- &- &- &89.6\\
                CLTD~\cite{xiong2024causality} & ECCV24 &98.6 &96.4 &89.3 &94.8\\ 
                Free Lunch~\cite{wang2024free} & ECCV24 &98.1 &94.1 &77.9 &90.0\\ 
                \hline
                \rowcolor{blue!8}Origins-T (\bftab{ours})  &- &\bftab{99.3} &\bftab{97.4} &\bftab{90.3} &\bftab{95.7}\\
				\hline
			\end{tabular}
			
		}
	\label{casiab}
\end{table}

\begin{table}[t]
\centering
\footnotesize
 	\caption{
        The performance comparisons on Gait3D are reported with Rank-1, Rank-5 accuracy and mAP (\%).
	}
        \setlength\tabcolsep{5pt}
		\resizebox{0.9999\linewidth}{!}{
			\begin{tabular}{c|c|ccc}
				\hline
				Method & Venue & Rank-1 & Rank-5 & mAP\\
				\hline
				GaitSet~\cite{chao2019gaitset} & AAAI19 &36.7 &58.3 &30.0 \\
				GaitPart~\cite{fan2020gaitpart} & CVPR20 &28.2 &47.6 &47.6 \\
				GaitGL~\cite{lin2021gait} & ICCV21 &29.7 &48.5 &22.3\\
				SMPLGait~\cite{zheng2022gait3d} & CVPR22 &46.3 &64.5 &37.2\\
                MTSGait~\cite{zheng2022gait} & MM22 &48.7 &67.1 &37.6\\
                QAGait~\cite{wang2024qagait} & AAAI24 &67.0 &81.5 &56.5\\
				GaitBase~\cite{fan2023opengait} & CVPR23 &64.6 &- &-\\
				DANet~\cite{ma2023dynamic} & CVPR23 &48.0 &69.7 &-\\
				GaitGCI~\cite{dou2023gaitgci} & CVPR23 &50.3 &68.5 &39.5\\
				DyGait~\cite{wang2023dygait} & ICCV23 &66.3 &80.8 &56.4\\
				HSTL~\cite{wang2023hierarchical} & ICCV23 &61.3 &76.3 &55.5\\
                VPNet~\cite{ma2024learning} & CVPR24 &75.4 &87.1 &-\\
                DeepGaitV2~\cite{11023027} & TPAMI25 &74.4 &88.0 &65.8\\
                CLTD~\cite{xiong2024causality} & ECCV24 &69.7 &85.2 &-\\ 
                GaitMoE~\cite{huang2024occluded} & ECCV24 &73.7 &- &66.2\\
                Free Lunch~\cite{wang2024free} & ECCV24 &70.1 &- &61.9\\ \hline
                \rowcolor{blue!8}Origins-M (\bftab{ours})  &- &\bftab{75.8} &\bftab{86.8} &\bftab{67.0}\\
				\hline
			\end{tabular}
			
		}
	\label{Gait3D}
\end{table}

\subsection{Results on in-the-wild Scenario}
We then validate the robustness of Origins against more complex and unknown covariates on in-the-wild datasets, SUSTech1K, Gait3D, GREW and CCGR-MINI.

\noindent \textbf{SUSTech1K.} As shown in Table.~\ref{SUSTech1K}, Origins-S demonstrates superior performance in more diverse and complex environments, surpassing state-of-the-art methods DeepGaitV2~\cite{11023027} by a significant margin 4.6\% in Overall Rank-1 accuracy. Notably, even facing the combination of various covariates, Origins achieves remarkable performance (\eg, 64.8\% Rank-1 accuracy in the Clothing benchmark), demonstrating its ability to preserve diverse semantic space during gait representation learning.

\noindent \textbf{Gait3D.} Origins-M achieves the SoTAs shown in Table.~\ref{Gait3D}, surpassing the latest appearance-based method CLTD~\cite{xiong2024causality} by 6.1\% in Rank-1 accuracy. Compared to the parameter-heavy GaitMoE~\cite{huang2024occluded}, it still maintain the improvements, indicating that Origins-M learns meaningful semantic space rather than noises.

\noindent \textbf{GREW.} Origins-L achieves competitive performance shown in Table.~\ref{GREW}, surpassing VPNet~\cite{ma2024learning} by 0.8\% Rank-1 accuracy. Origins-L adopts Free Lunch~\cite{wang2024free} (\ie, logits as gait representations) to achieve more stable results without introducing any additional computational complexity. 

\noindent \textbf{CCGR-MINI.} As shown in Table.~\ref{CCGR}, Origins-M outperforms the previous SoTA DeepGaitV2 by 2.1\% Rank-1 accuracy, which further reveals the robustness.

\begin{table}[t]
\centering
\footnotesize
 	\caption{
        The performance comparisons on GREW are reported with Rank-1, Rank-5 and Rank-10 accuracy (\%).
	}
        \setlength\tabcolsep{4pt}
		\resizebox{0.9999\linewidth}{!}{
			\begin{tabular}{c|c|ccc}
				\hline
				Method & Venue & Rank-1 & Rank-5 & Rank-10\\
				\hline
				GaitSet~\cite{chao2019gaitset} & AAAI19 &46.3 &63.6 &70.3\\
				GaitPart~\cite{fan2020gaitpart} & CVPR20 &44.0 &60.7 &67.3\\
				GaitGL~\cite{lin2021gait} & ICCV21 &47.3 &63.6 &-\\
                MTSGait~\cite{zheng2022gait} & MM22 &55.3 &71.3 &76.9\\
                QAGait~\cite{wang2024qagait} & AAAI24 &59.1 &74.0 &79.2\\
				GaitBase~\cite{fan2023opengait} & CVPR23 &60.1 &- &-\\
				GaitGCI~\cite{dou2023gaitgci} & CVPR23 &68.5 &80.8 &84.9\\
				DyGait~\cite{wang2023dygait} & ICCV23 &71.4 &83.2 &86.8\\
				HSTL~\cite{wang2023hierarchical} & ICCV23 &62.7 &76.6 &81.3\\
                VPNet~\cite{ma2024learning} & CVPR24 &80.0 &89.4 &-\\
                DeepGaitV2~\cite{11023027} & TPAMI25 &77.7 &88.9 &91.8\\
                CLTD~\cite{xiong2024causality} & ECCV24 &78.0 &87.8 &-\\ 
                GaitMoE~\cite{huang2024occluded} & ECCV24 &79.6 &89.1 &-\\
                Free Lunch~\cite{wang2024free} & ECCV24 &65.5 &78.7 &83.3\\ 
                \hline
                \rowcolor{blue!8}Origins-L (\bftab{ours})  &- &\bftab{80.8} &\bftab{89.6} &\bftab{92.1}\\
				\hline
			\end{tabular}
			
		}
	\label{GREW}
\end{table}

\begin{table}[t]
\centering
\footnotesize
 	\caption{
        The performance comparisons on CCGR-MINI are reported with Rank-1 accuracy, mAP and mINP (\%).
	}
        \setlength\tabcolsep{6pt}
		\resizebox{0.9999\linewidth}{!}{
			\begin{tabular}{c|c|ccc}
				\hline
				Method & Venue & Rank-1 & mAP & mINP\\
				\hline
				GaitSet~\cite{chao2019gaitset} & AAAI19 &13.77	&15.39	&5.75\\
				GaitPart~\cite{fan2020gaitpart} & CVPR20	&8.02	&10.12	&3.52\\
				GaitGL~\cite{lin2021gait} & ICCV21&17.51	&18.12	&6.85\\
				GaitBase~\cite{fan2023opengait} & CVPR23&26.99	&24.89	&9.72\\
                DeepGaitV2~\cite{11023027} & TPAMI25&39.37	&36.01	&16.77\\
                \hline
                \rowcolor{blue!8}Origins-M (\bftab{ours})	&- &\bftab{41.45}	&\bftab{38.31}	&\bftab{24.71}\\
				\hline
			\end{tabular}
			
		}
	\label{CCGR}
\end{table}

\subsection{Ablation Study}

We first validate the core modules of Origins on Gait3D. Then, we analyze the number of timestep templates on Gait3D and CASIA-B. Finally, we perform visualization to better understand the mechanisms of Origins.

\noindent \textbf{The core modules of Origins.} As shown in Table.~\ref{ab_gait3d}, Origins achieved significant improvments compared to the Baseline with 2.4\% higher Rank-1 accuracy on Gait3D. Specifically, the combination of Transferring Unified Template and Diffusing Timestep Templates enables further improving performance, proving the effectiveness of unified gait generative and representation learning.
Additionally, we find that the number of multi-heads $\mathcal{Z}$ in attention mechanism is set to 16 for better performance shown in Table.~\ref{ab_gait3d}.

\noindent \textbf{The number of timestep templates.} As shown in Table.~\ref{ab_gait3d_casiab}, the number of timestep templates represents a trade-off. For a large gait database with complex covariates, having fewer timestep templates may result in a narrow semantic space, potentially affecting the expression of gait representations. Conversely, more timestep templates could lead to an overwhelming semantic space that introduces noise and confusion into the representations.
For the smaller datasets, the number of timestep templates has less impact, which is make sense as gait representation learning already enable to capture sufficient gait patterns.

\begin{figure}[tbp]
    \centering
    \begin{subfigure}[b]{0.23\textwidth}
        \centering
        \includegraphics[width=\textwidth]{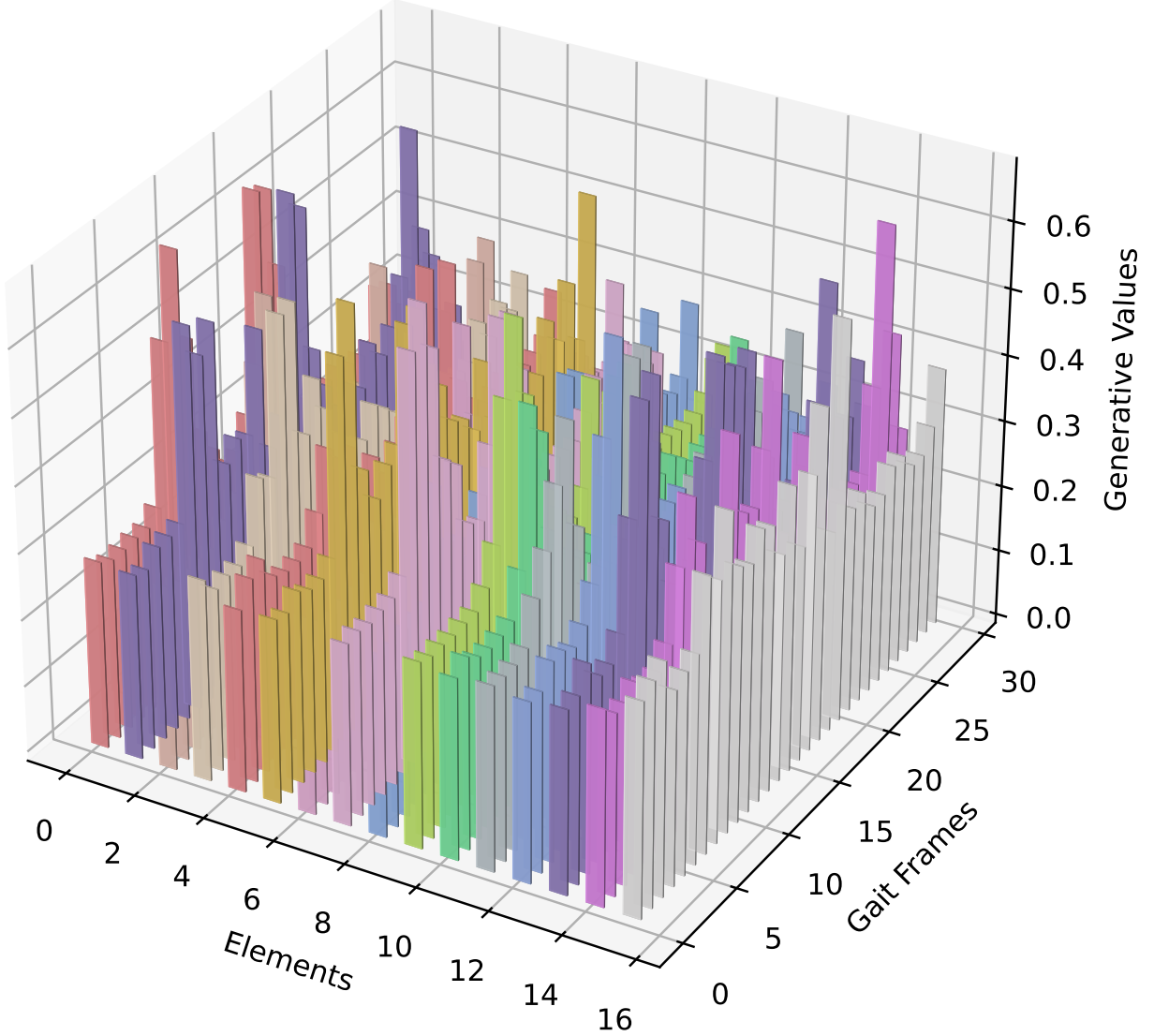}
        \caption{ID 77 with the timestep template}
        \label{fig:subfig3}
    \end{subfigure}
    \hfill
    \begin{subfigure}[b]{0.23\textwidth}
        \centering
        \includegraphics[width=\textwidth]{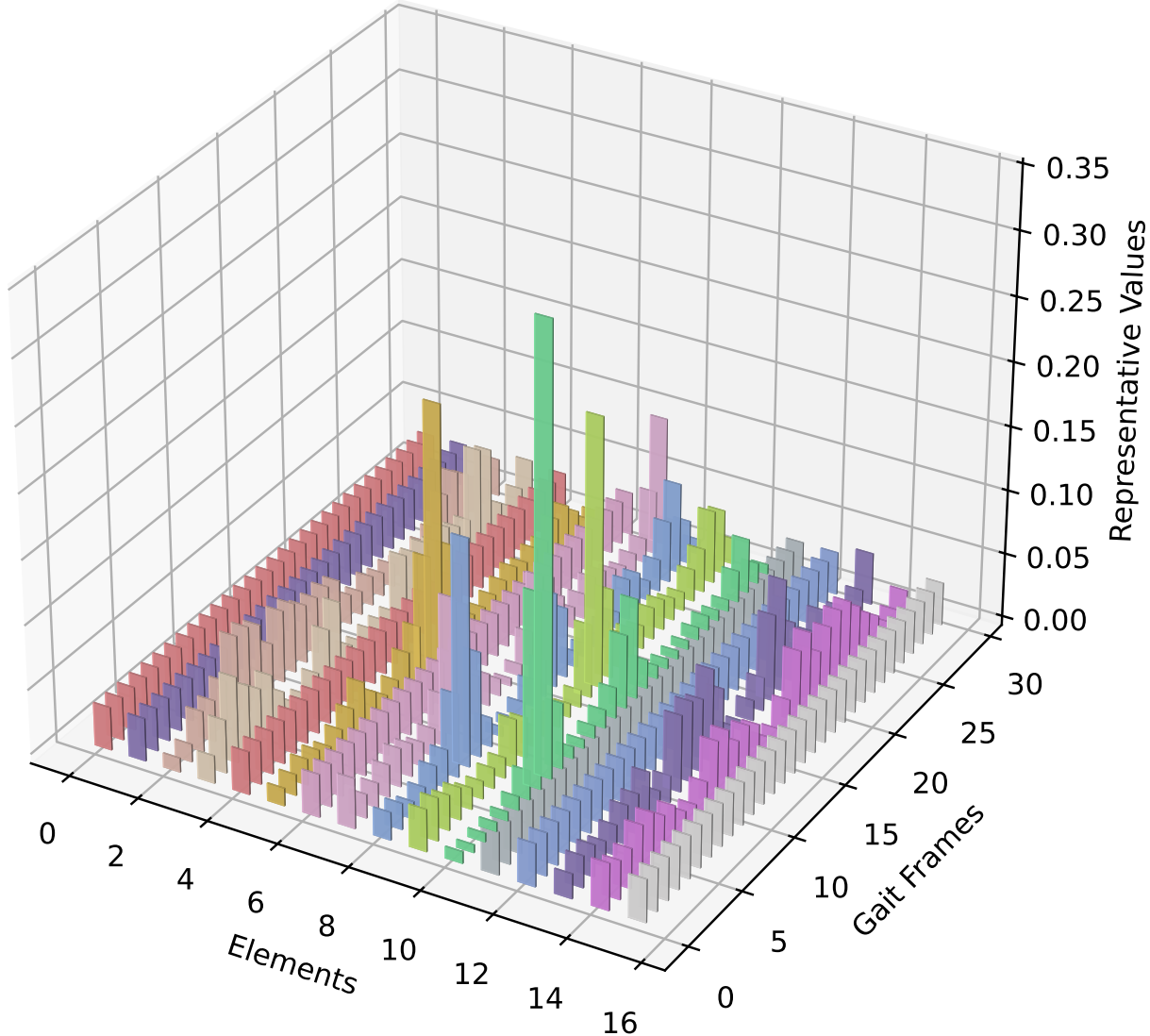}
        \caption{ID 77 with the unified template}
        \label{fig:subfig4}
    \end{subfigure}

    \vspace{0.5cm} 
    \begin{subfigure}[b]{0.23\textwidth}
        \centering
        \includegraphics[width=\textwidth]{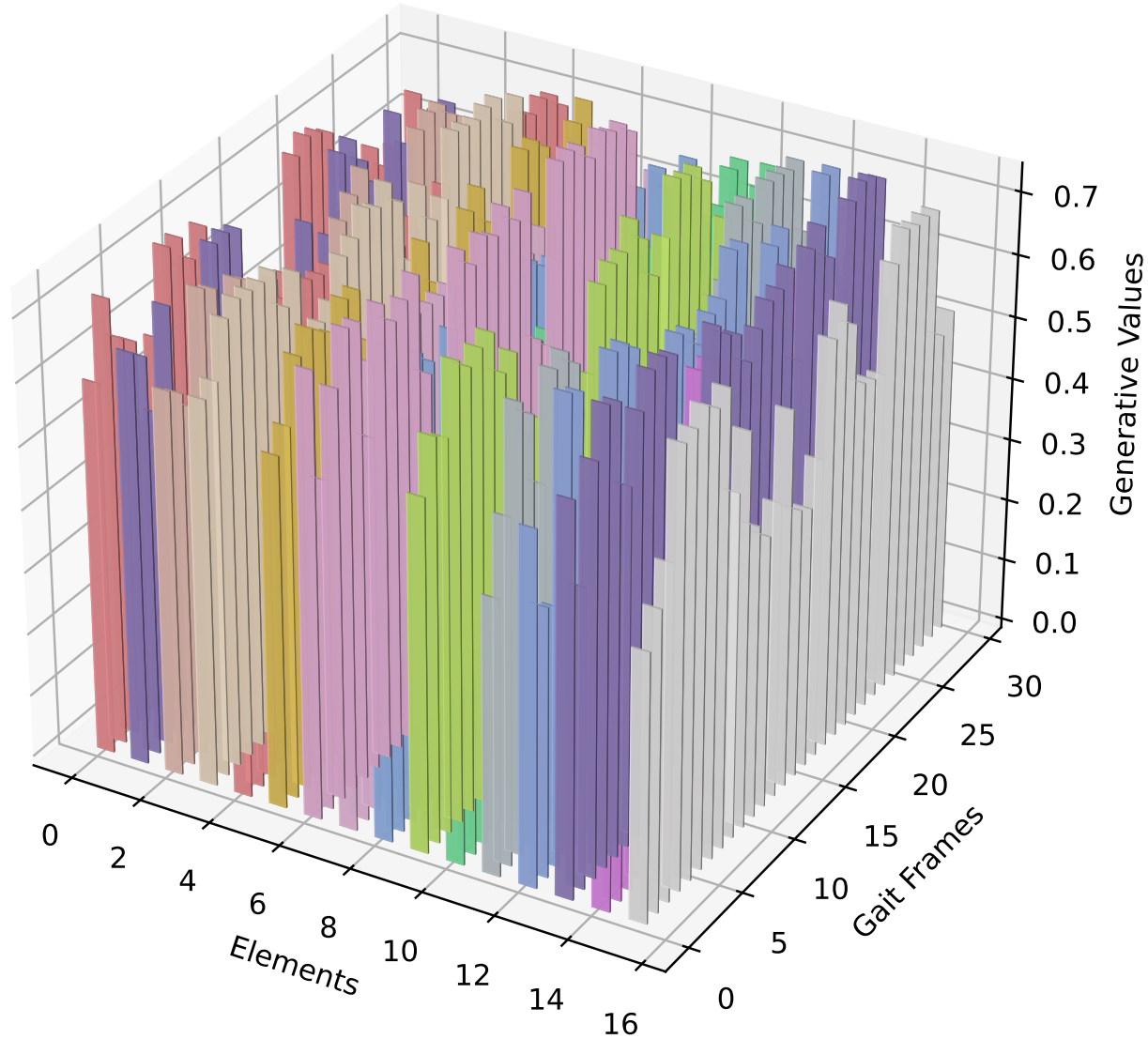}
        \caption{ID 75 with the timestep template}
        \label{fig:subfig7}
    \end{subfigure}
    \hfill
    \begin{subfigure}[b]{0.23\textwidth}
        \centering
        \includegraphics[width=\textwidth]{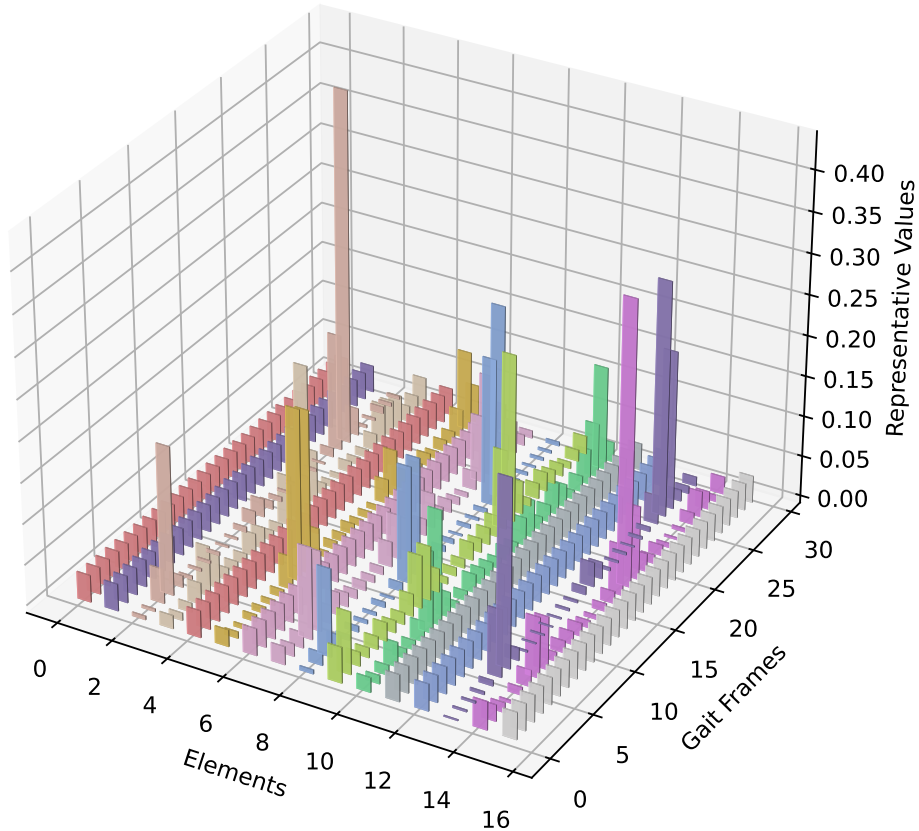}
        \caption{ID 75 with the unified template}
        \label{fig:subfig8}
    \end{subfigure}
    \caption{The ``elements'' axe is $\mathcal Z$ that is the number of multi-heads in the template. The visualization shows that different subjects exhibit differences in the template.}
    \label{visual}
\end{figure}

\begin{table}[t]
\centering
 	\caption{
        The core module analysis on Gait3D.
	}
        \setlength\tabcolsep{5.5pt}
		\resizebox{0.9999\linewidth}{!}{%
			\begin{tabular}{c|ccc}
				\hline
				\multirow{2}{*}{Method} & \multicolumn{3}{c}{Gait3D}\\
				\cline{2-4}
				& Rank-1 & Rank-5 &mAP\\
				\hline
                Origins-M &75.8 &86.8 &67.0\\ 
                \hline
                \rowcolor{blue!8} \multicolumn{4}{c}{The analysis on the core modules}\\ \hline
                 -w/o Origins &73.6 &86.9 &65.1\\
                 - w/o Transferring Unified Template	&74.4 &87.9 &67.1\\ 
                 - w/o Diffusing Timestep Templates	&75.3	&86.4 &66.3\\
				\hline
               \rowcolor{blue!8} \multicolumn{4}{c}{The analysis on the number of multi-heads}\\ 
               \hline
               $\mathcal{Z}$ = 8 &74.6 &86.6 &66.7\\
               $\mathcal{Z}$ = 16 &75.8 &86.8 &67.0\\
               $\mathcal{Z}$ = 32 &75.2 &86.4 &66.1 \\
               \hline
			\end{tabular}
			
		}
	\label{ab_gait3d}
\end{table}

\begin{table}[t]
\centering
 	\caption{
        The timestep template analysis on Gait3D and CASIA-B.
	}
        \setlength\tabcolsep{4pt}
		\resizebox{0.9999\linewidth}{!}{%
			\begin{tabular}{c|ccc|cccc}
				\hline
				\multirow{2}{*}{Method} & \multicolumn{3}{c}{Gait3D} &\multicolumn{4}{c}{CASIA-B}\\
				\cline{2-8}
				& Rank-1 & Rank-5 &mAP &NM &BG &CL &Mean\\
				\hline
                Origins-M &75.8 &86.8 &67.0 &99.3 &97.4 &90.3 &95.7\\ 
                Baseline &73.6 &86.9 &65.1 &98.2 &95.8 &84.4 &92.8\\
                \hline
                \rowcolor{blue!8} \multicolumn{8}{c}{ The analysis on the number of timestep templates}\\ \hline
                 $\mathcal{N} = 8$ &75.3 &87.0 &66.5 &99.1 &97.3 &90.4 &95.6\\ 
                 $\mathcal{N} = 16$ &75.8 &86.8 &67.0 &99.3 &97.4 &90.3 &95.7\\ 
                 $\mathcal{N} = 32$ &73.7 &86.3 &65.4 &99.2 &97.3 &90.0 &95.5\\ 

				\hline
			\end{tabular}
			
		}
	\label{ab_gait3d_casiab}
\end{table}

\noindent \textbf{The visualization of the unified Template.} As shown in (a, c) of Fig.~\ref{visual}, we perform the generative differences to better understand the mechanism of Diffusing Timestep Templates. Specifically, we select the 8th timestep template to generate gait representations for different IDs. It can be observed that different IDs impact the signal from the timestep template. However, they can be linearly combined by these elements (\ie, the multi-heads $\mathcal{Z}$), which means they are regularized to a consistent space.
As shown in (b, d) of Fig.~\ref{visual}, we perform the discriminative differences to better understand the mechanism. Transferring Unified Template aims to compresses each gait sequence into one token as completely and effectively as possible.
It can be observed that not all information within a gait sequence effectively contribute to recognition. Instead, the unified template captures accurate gait differences in the multi-heads $\mathcal{Z}$ attention mechanism for recognition. We also observe the gait periodicity in some heads, highlighting the impact of fine-grained body motion. 

%% file: sec/5_con.tex
\section{Conclusion and Limitations}

In this work, we propose Origins with generative capabilities, regularizing gait representations within a consistent and diverse semantic space and addressing semantic inconsistency and uniformity in complex scenarios. Origins learns a unified template through diffusing timestep templates for gait generative learning, addressing the convergence difficulty, and meanwhile transfers the unified template for gait representation learning, capturing accurate gait differences. Extensive experiments demonstrate that Origins performs unified generative and representation learning, achieving superior performance.

\noindent \textbf{Limitations.} While diffusion models have been widely adopted to generate new samples at the pixel level, enhancing the usability, Origins performs generative learning in the representation space. 
In the future, we will explore generating new gait samples to further enhance general representations, such as in pretraining paradigms.

\section*{Acknowledgement}
\label{sec:acknowledgement}
This work is jointly supported by National Natural Science Foundation of China (62276025, 62206022, 62476027) and the Fundamental Research Funds for the Central Universities (2253200026).